%% file: acl2020.tex
\documentclass[11pt,a4paper]{article}
\usepackage[hyperref]{acl2020}
\usepackage{times}
\usepackage{latexsym}

\usepackage{microtype}

\aclfinalcopy 

\usepackage[normalem]{ulem}
\usepackage{multirow}
\usepackage{makecell}
\usepackage{soul}
\usepackage{enumitem}
\usepackage{adjustbox}
\usepackage{amsmath}
\usepackage{amssymb}
\usepackage{array}
\usepackage{graphicx}
\usepackage{comment}
\usepackage{booktabs}
\usepackage{xcolor,colortbl} 
\usepackage{algorithm}
\usepackage{algorithmic}

\usepackage{bbm}
\usepackage{tabularx}

\newcommand{\real}[1]{{\mathbb{R}^{{#1}}}}

\title{Explicit Memory Tracker with Coarse-to-Fine Reasoning for Conversational Machine Reading}

\author{\makecell{
Yifan Gao$^\dag$\thanks{~~~This work was mostly done when Yifan Gao was an intern at Salesforce Research Asia, Singapore.}, Chien-Sheng Wu$^\ddag$, Shafiq Joty$^{\ddag}$, Caiming Xiong$^\ddag$, Richard Socher$^\ddag$,\\
Irwin King$^\dag$, Michael R. Lyu$^\dag$, and Steven C.H. Hoi$^\ddag$}
\\
	{$^\dag$ The Chinese University of Hong Kong}\\
	{$^\ddag$ Salesforce Research}\\
    \tt{ $^\dag$\{yfgao,king,lyu\}@cse.cuhk.edu.hk}\\
	\tt{ $^\ddag$\{cswu,sjoty,cxiong,rsocher,shoi\}@salesforce.com}
}

\date{}

\begin{document}
\maketitle
\begin{abstract}
The goal of conversational machine reading is to answer user questions given a knowledge base text which may require asking clarification questions. Existing approaches are limited in their decision making due to struggles in extracting question-related rules and reasoning about them. 
In this paper, we present a new framework of conversational machine reading that comprises a novel \textbf{E}xplicit \textbf{M}emory \textbf{T}racker (EMT) to track whether conditions listed in the rule text have already been satisfied to make a decision.
Moreover, our framework generates clarification questions by adopting a coarse-to-fine reasoning strategy, utilizing sentence-level entailment scores to weight token-level distributions. 
On the ShARC  benchmark (blind, held-out) testset, EMT achieves new state-of-the-art results of 74.6\% micro-averaged decision accuracy and 49.5 BLEU4. 
We also show that EMT is more interpretable by visualizing the entailment-oriented reasoning process as the conversation flows.
Code and models are released at \url{https://github.com/Yifan-Gao/explicit_memory_tracker}.

\end{abstract}

\section{Introduction}
In conversational machine reading (CMR), machines can take the initiative to ask users questions that help to solve their problems, instead of jumping into a conclusion hurriedly~\cite{saeidi-etal-2018-interpretation}.
In this case, machines need to understand the knowledge base (KB) text, evaluate and keep track of the user scenario, ask clarification questions, and then make a final decision.
This interactive behavior between users and machines has gained more attention recently because in practice users are unaware of the KB text, thus they cannot provide all the information needed in a single turn.

\begin{figure}[t!]
\centering
\includegraphics[width=1.0\columnwidth]{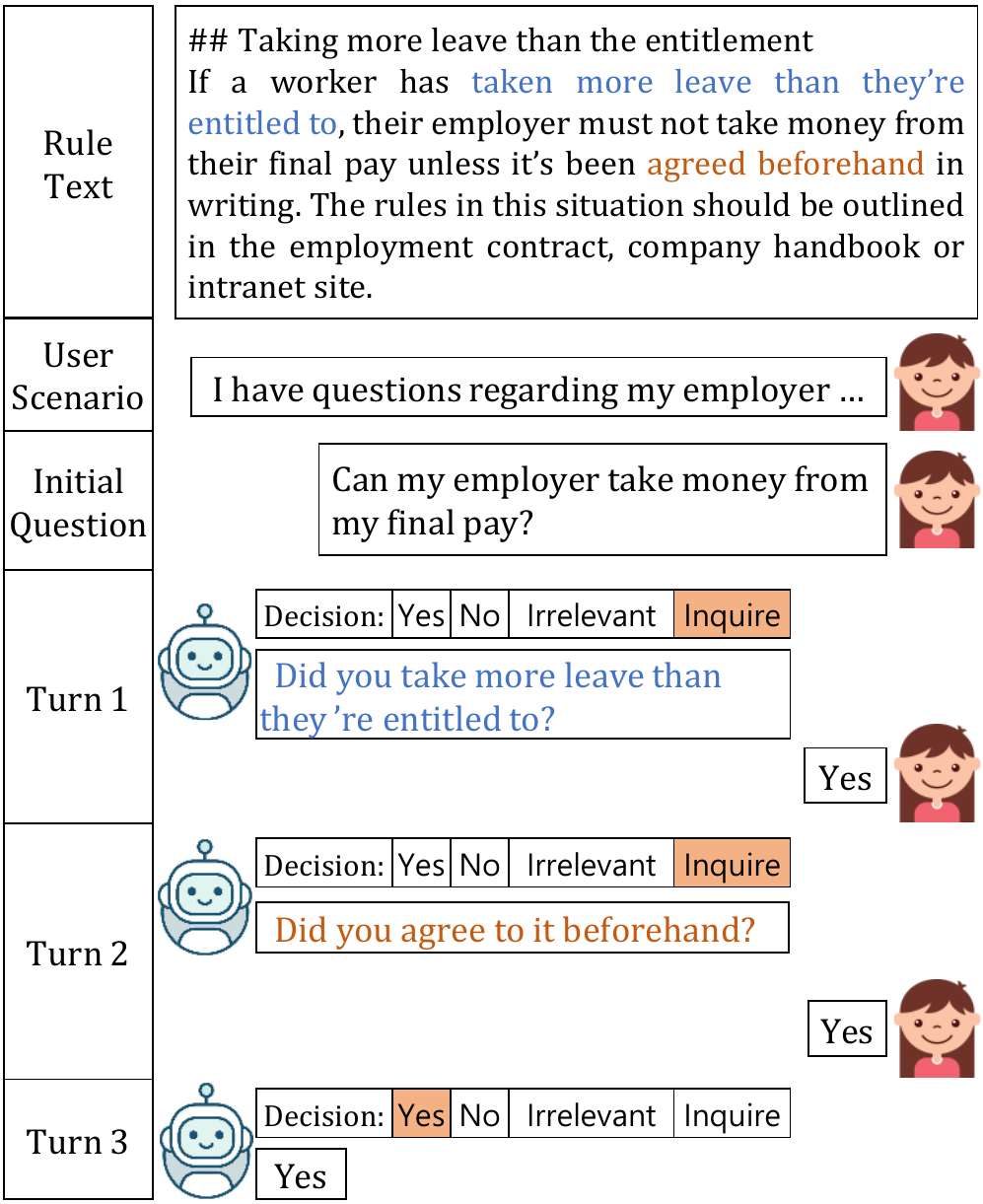}
\caption{Example of Conversational Machine Reading tasks from the ShARC dataset \cite{saeidi-etal-2018-interpretation}. At each turn, given the rule text, a user scenario, an initial user question, and previous interactions, a machine can give a certain final answer such as \texttt{Yes} or \texttt{No} to the initial question. If the machine cannot give a certain answer because of missing information from the user, it will ask a clarification question to fill in the information gap.
Clarification questions and their corresponding rules are marked in the same colors.
}
\vspace{-0.1in}
\label{fig:example}
\end{figure}

For instance, consider the example in Figure \ref{fig:example} taken from the ShARC dataset for CMR \cite{saeidi-etal-2018-interpretation}. A user posts her scenario and asks a question on whether her employer can take money from her final pay. Since she does not know the relevant rule text, the provided scenario and the initial question(s) from her are often too underspecified for a machine to make a certain decision. Therefore, a machine has to read the rule text and ask a series of clarification questions until it can conclude the conversation with a certain answer.

Most existing approaches~\cite{zhong-zettlemoyer-2019-e3,Sharma2019NeuralCQ} formalize the CMR problem into two sub-tasks. The first is to make a decision among \texttt{Yes}, \texttt{No}, \texttt{Irrelevant}, and \texttt{Inquire} at each dialog turn given a rule text, a user scenario, an initial question and the current dialog history. If one of \texttt{Yes}, \texttt{No}, or \texttt{Irrelevant} is selected, it implies that a final decision (\texttt{Yes}/\texttt{No}) can be made in response to the user's initial question, or stating the user's initial question is unanswerable (\texttt{Irrelevant}) according to the rule text. If the decision at the current turn is \texttt{Inquire}, it will then trigger the second task for follow-up question generation, which extracts an underspecified rule span from the rule text and generates a follow-up question accordingly.

However, there are two main drawbacks to the existing methods. First, with respect to the reasoning of the  
rule text, existing methods do not explicitly track whether a condition listed in the rule has already been satisfied as the conversation flows so that it can make a better decision. Second, with respect to the extraction of question-related rules, it is difficult in the current approach to extract the most relevant text span to generate the next question. 
For example, the state-of-the-art E${}^3$ model~\cite{zhong-zettlemoyer-2019-e3} has only 60.6\% F1 for question-related span extraction.

To address these issues, we propose a new framework of conversational machine reading with a novel \textbf{E}xplicit \textbf{M}emory \textbf{T}racker (EMT), which explicitly tracks each rule sentence to make decisions and generate follow-up questions. 
Specifically, EMT first segments the rule text into several rule sentences and allocates them into its memory.
Then the initial question, user scenario, and dialog history are fed into EMT sequentially to update each memory module separately.
At each dialog turn, EMT predicts the entailment states (satisfaction or not) for every rule sentence, and makes a decision based on the current memory status.
If the decision is \texttt{Inquire}, EMT extracts a rule span to generate a follow-up question by adopting a coarse-to-fine reasoning strategy (i.e., weighting token-level span distributions with its sentence-level entailment scores).
Compared to previous methods which only consider entailment-oriented reasoning for decision making or follow-up question generation, EMT utilizes its updated memory modules to reason out these two tasks in a unified manner.

We compare EMT with the existing approaches on the ShARC dataset~\cite{saeidi-etal-2018-interpretation}. Our results show that explicitly tracking rules with external memories boosts both the decision accuracy and the quality of generated follow-up questions.
In particular, EMT outperforms the previous best model E${}^3$ by 1.3 in macro-averaged decision accuracy and 10.8 in BLEU4 for follow-up question generation.
In addition to the performance improvement, EMT yields interpretability by explicitly tracking rules, which is visualized to show the entailment-oriented reasoning process of our model.

\begin{figure*}[t!]
\centering
\includegraphics[width=1.0\textwidth]{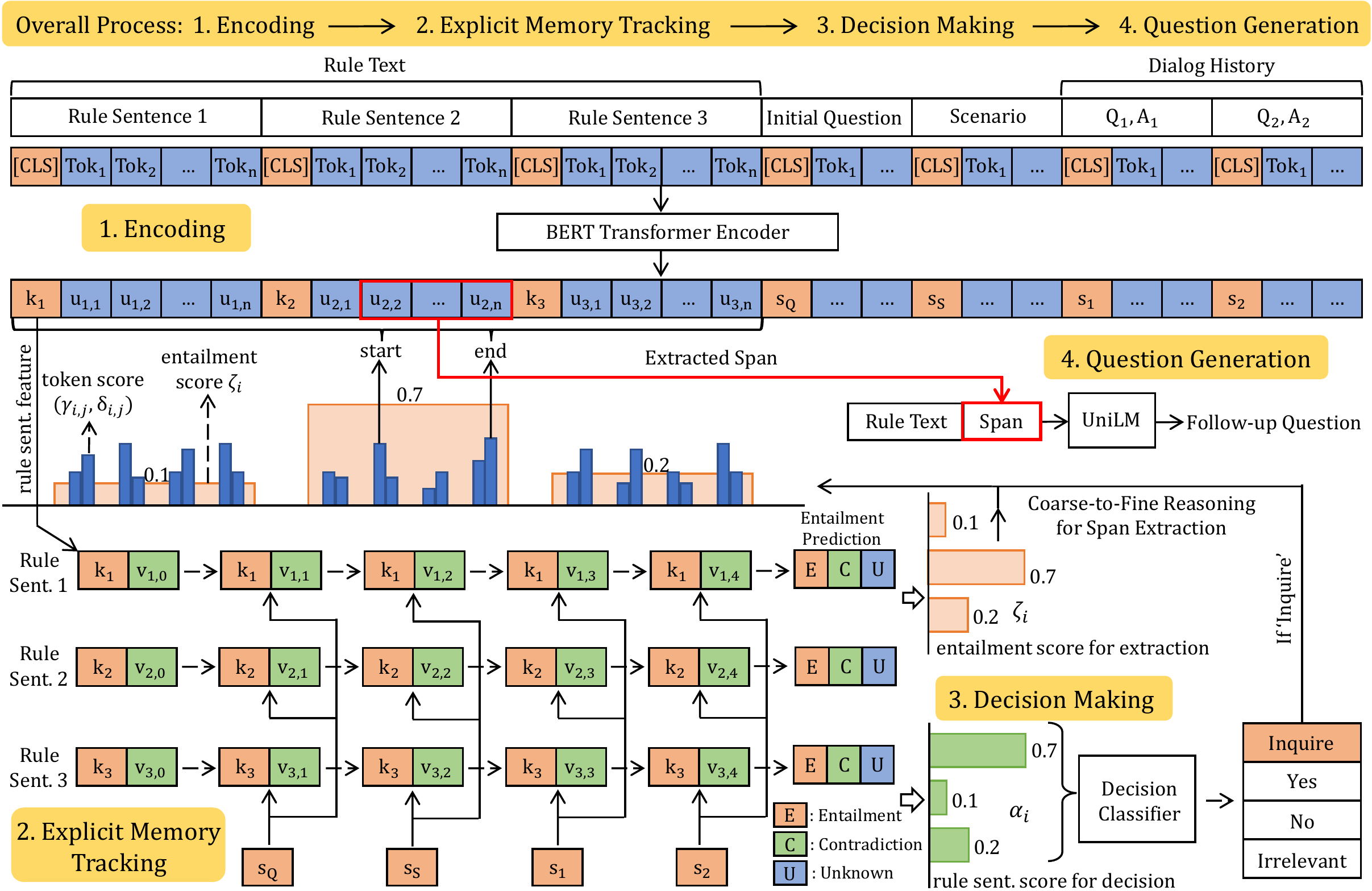}
\caption{The Explicit Memory Tracker with Coarse-to-Fine Reasoning for Conversational Machine Reading (CMR). The CMR process includes (1) BERT encoding, (2) Explicit Memory Tracking for entailment state of each rule sentence, (3) Decision Making on updated entailment states of all rule sentences, (4) Question Generation via span extraction with coarse-to-fine reasoning and question rephrasing of the extracted span. \textit{(Best viewed in color)}}
\label{fig:model}
\end{figure*}

\section{Method}

As illustrated in Figure \ref{fig:model}, our proposed method consists of the following four main modules. 
\begin{itemize}
\vspace{-0.1in}
\item[(1)] The \emph{Encoding} module uses BERT \cite{devlin-etal-2019-bert} to encode the concatenation of the rule text, initial question, scenario and dialog history into contextualized representations.\vspace{-0.1in}
\item[(2)] The \emph{Explicit Memory Tracking} module sequentially reads the initial question, user scenario, multi-turn dialog history, and updates the entailment state of each rule sentence.\vspace{-0.1in}
\item[(3)] The \emph{Decision Making} module does entailment-oriented reasoning based on the updated states of rule sentences and makes a decision among \texttt{Yes}, \texttt{No}, \texttt{Irrelevant}, and  \texttt{Inquire}.\vspace{-0.1in}
\item[(4)] If the decision is \texttt{Inquire}, the \emph{Question Generation} module is activated, which reuses the updated states of rule sentences to identify the underspecified rule sentence and extract the most informative span within it in a coarse-to-fine manner. Then it rephrases the extracted span into a well-formed follow-up question.
\end{itemize}

\subsection{Encoding}
Let $x_R$, $x_Q$, $x_S$, $[x_{H,1}, x_{H,2}, ...,  x_{H,P}]$ denote the input of rule text, initial question, user scenario, and $P$ turns of dialog history, each of which is a sequence of tokens.
We first split the rule text $x_R$ into several rule sentences $[x_{R,1}, x_{R,2}, ..., x_{R,M}]$ according to sentence boundary or bullet points, insert \texttt{[CLS]} tokens at the start of each sentence, and concatenate them into one sequence:
\\ \noindent[\texttt{[CLS]}, $x_{R,1}$; ... ; \texttt{[CLS]}, $x_{R,M}$; \texttt{[CLS]}, $x_Q$; \texttt{[CLS]}, $x_S$; \texttt{[CLS]}, $x_{H,1}$; ... ; \texttt{[CLS]}, $x_{H,P}$].
Then we use BERT \cite{devlin-etal-2019-bert}, a pretrained transformer encoder \cite{transformer} to encode the sequence into a sequence of vectors with the same length. 
We treat each \texttt{[CLS]} representation as feature representation of the  sentence that follows it. In this way, we receive both token-level representation and sentence-level representation for each sentence.
We denote sentence-level representation of the rule sentences as $\mathbf k_1, ..., \mathbf k_M$ and their token-level representation as $[(\mathbf u_{1,1}, ..., \mathbf u_{1,n_1}), ..., (\mathbf u_{M,1}, ..., \mathbf u_{M,n_M})]$, where $n_i$ is number of tokens for rule sentence $i$. 
Similarly, we denote the sentence-level representation of the initial question, user scenario, and $P$ turns of dialog history as $\mathbf s_Q$, $\mathbf s_S$, and $\mathbf s_1, ..., \mathbf s_P$, respectively.
All these vectorized representations are of $d$ dimensions (768 for BERT-base).

\subsection{Explicit Memory Tracking}\label{sec:emt}
Given the rule sentences $\mathbf k_1, ..., \mathbf k_M$ and the user provided information including the initial question $\mathbf s_Q$, scenario  $\mathbf s_S$, and $P$ turns of dialog history $\mathbf s_1, ..., \mathbf s_P$, our goal is to find implications between the rule sentences and the user provided information.
Inspired by Recurrent Entity Network \cite{DBLP:conf/iclr/HenaffWSBL17} which tracks the world state given a sequence of textual statements, we propose the \textbf{E}xplicit \textbf{M}emory \textbf{T}racker (EMT), a gated recurrent memory-augmented neural network which explicitly tracks the states of rule sentences by sequentially reading the user provided information.

As shown in Figure \ref{fig:model}, EMT explicitly takes rule sentences $\mathbf k_1, ..., \mathbf k_M$ as keys, and assigns a state $\mathbf v_i$ to each key to save the most updated entailment information (whether this rule has been entailed from the user provided information).
Each value state $\mathbf v_i$ is initialized with the same value of its corresponding rule sentence:  $\mathbf v_{i,0} = \mathbf k_i$.
Then EMT sequentially reads user provided information $\mathbf s_Q, \mathbf s_S, \mathbf s_1, ..., \mathbf s_P$. 
At time step $t$, the value state $\mathbf v_{i,t}$ for $i$-th rule sentence is updated by incorporating the user provided information $\mathbf s_t \in \{\mathbf s_Q, \mathbf s_S, \mathbf s_1, ..., \mathbf s_P\}$,
\begin{align}
    {\tilde{\mathbf{v}}}_{i,t} &= \text{ReLU} (\mathbf W_k \mathbf k_i + \mathbf W_v \mathbf v_{i,t} + \mathbf W_s \mathbf s_t), \\
    g_i &= \sigma(\mathbf s_t^\top \mathbf k_i + \mathbf s_t^\top \mathbf v_{i,t}) \in [0,1], \label{eqn:gate} \\
    \mathbf v_{i,t} &= \mathbf v_{i,t} + g_i \odot {\tilde{\mathbf{v}}}_{i,t} \in \real{d}, 
    \mathbf v_{i,t} = \frac{\mathbf v_{i,t}}{\Vert \mathbf v_{i,t} \Vert}
\end{align}
where $\mathbf W_k, \mathbf W_v, \mathbf W_s \in \real{d \times d}$, $\sigma$ represents a sigmoid function, and $\odot$ is scalar product.
As the user background input $\mathbf s_t$ may only be relevant to parts of the rule sentences, the gating function in Equation \ref{eqn:gate} matches $\mathbf s_t$ to the memory. Then EMT updates state $\mathbf v_{i,t}$ only in a gated manner. Finally, the normalization allows EMT to forget previous information, if necessary.
After EMT sequentially reads all user provided information (the initial question, scenario, and $P$ turns of history dialog) and finishes entailment-oriented reasoning, keys and final states of rule sentences are denoted as $(\mathbf k_1,  \mathbf v_{1}), ..., (\mathbf k_M, \mathbf v_{M})$, which will be used in the decision making module (Section \ref{sec:clf}) and question generation module (Section \ref{sec:qg}).

The key difference between our Explicit Memory Tracker and Recurrent Entity Network (REN) \cite{DBLP:conf/iclr/HenaffWSBL17} is that each key $\mathbf k_i$ in our case has an explicit meaning (the corresponding rule sentence) and thus it changes according to different rule texts while in REN, the underlined meaning of keys are learned through training and they are fixed throughout all textual inputs.
Moreover, the number of keys is dynamic in our case (according to the number of sentences parsed from the rule text) while that is predefined in REN.

\subsection{Decision Making} \label{sec:clf}
Based on the most up-to-date key-value states of rule sentences $(\mathbf k_1,  \mathbf v_{1}), ..., (\mathbf k_M, \mathbf v_{M})$ from the EMT, the decision making module predicts a decision among \texttt{Yes, No, Irrelevant}, and \texttt{Inquire}.
First, we use self-attention to compute a summary vector $\mathbf c$ for the overall state: 
\begin{align}
    \alpha_i &= \mathbf w_{\alpha}^\top [\mathbf k_i; \mathbf v_{i}] + b_\alpha \in \real{1} \label{eqn:alpha} \\
    \tilde{\alpha}_i &= \text{softmax}(\mathbf{\alpha})_i \in [0,1]  \\
    \mathbf c &= \sum_i \tilde{\alpha_i} [\mathbf k_i; \mathbf v_{i}] \in \real{d}
\end{align}
where $[\mathbf k_i; \mathbf v_{i}]$ denotes the concatenation of the vectors $\mathbf k_i$ and $\mathbf v_{i}$, and $\alpha_i$ is the attention weight for the rule sentence $\mathbf{k}_i$ that determines the likelihood that $\mathbf{k}_i$ is entailed from the user provided information. 
\\\noindent Then the final decision is made through a linear transformation of the summary vector $\mathbf c$:
\begin{align}
    \mathbf z &= \mathbf W_z \mathbf c + \mathbf b_z \in \real{4}
\end{align}
where $\mathbf z \in \real{4}$ contains the model's score for all four possible classes.
Let $l$ indicate the correct decision, the decision making module is trained with the following cross entropy loss:
\begin{align}\label{eqn:l_dec}
    \mathcal{L}_{\text{dec}} = -\log~\text{softmax}(\mathbf z)_l
\end{align}

In order to explicitly track whether a condition listed in the rule has already been satisfied or not, we add a subtask to predict the entailment states for each rule sentence. The possible entailment labels are \texttt{Entailment}, \texttt{Contradiction} and \texttt{Unknown}; details of acquiring such labels are described in Section \ref{sec:ex_setup}. With this intermediate supervision, the model can make better decisions based on the correct entailment state of each rule sentence.
The entailment prediction is made through a linear transformation of the most up-to-date key-value state $[\mathbf k_i; \mathbf v_{i}]$ from the EMT module:
\begin{align}\label{eqn:score_entail}
    \mathbf e_i &= \mathbf W_e [\mathbf k_i; \mathbf v_{i}] + \mathbf b_e \in \real{3}
\end{align}
where $\mathbf e_i \in \real{3}$ contains scores of three entailment states [$\beta_{\text{entailment},i}$, $\beta_{\text{contradiction},i}$, $\beta_{\text{unknown},i}$] for the $i$-th rule sentence.
Let $r$ indicate the correct entailment state. The entailment prediction subtask is trained with the following cross entropy loss, normalized by the number of rule sentences $M$:
\begin{align}\label{eqn:l_entail}
    \mathcal{L}_{\text{entail}} = - \frac{1}{M}\sum_{i=1}^M ~ \log~\text{softmax}(\mathbf e_i)_r
\end{align}

\subsection{Follow-up Question Generation} \label{sec:qg}
When the decision making module predicts \texttt{Inquire}, a follow-up question is required for further clarification from the user.
In the same spirit of previous studies \cite{zhong-zettlemoyer-2019-e3, Sharma2019NeuralCQ}, we decompose this problem into two stages. First, we extract a span inside the rule text which contains the underspecified user information (we name it as \texttt{underspecified span} hereafter). Second, we rephrase the extracted underspecified span into a follow-up question. We propose a coarse-to-fine approach to extract the underspecified span for the first stage, and finetune the pretrained language model UniLM \cite{unilm} for the follow-up question rephrasing, as we describe below.

\paragraph{Coarse-to-Fine Reasoning for Underspecified Span Extraction.}
 \citet{zhong-zettlemoyer-2019-e3} extract the underspecified span by extracting several spans and retrieving the most likely one. The disadvantage of their approach is that extracting multiple rule spans is a challenging task, and it will propagate errors to the retrieval stage. Instead of extracting multiple spans from the rule text, we propose a coarse-to-fine reasoning approach to directly identify the underspecified span. For this, we reuse the \texttt{Unknown} scores $\beta_{\text{unknown},i}$ from the entailment prediction subtask (Eqn. \ref{eqn:score_entail}), and normalize it (over the rule sentences) with a softmax to determine how likely that the $i$-th rule sentence contains the underspecified span: 
\begin{align}\label{eqn:entail_sent_score}
    \zeta_i &= \text{softmax}(\mathbf{\beta}_{\text{unknown}})_i \in [0,1]
\end{align}

Knowing how likely a rule sentence is underspecified greatly reduces the difficulty to extract the underspecified span within it. 
We adopt a soft selection approach to modulate span extraction (i.e., predicting the start and end points of a span) score by rule sentence identification score $\zeta_i$.
We follow the BERTQA approach \cite{devlin-etal-2019-bert} to learn a start vector $\mathbf w_s \in \real{d}$ and an end vector $\mathbf w_e \in \real{d}$ to locate the start and end positions from the whole rule text.
The probability of $j$-th word in $i$-th rule sentence $\mathbf u_{i,j}$ being the start/end of the span is computed as a dot product between $\mathbf w_s$ and $\mathbf u_{i,j}$, modulated by its rule sentence score $\zeta_i$:
\begin{align}\label{eqn:modulate}
    \gamma_{i,j} = \mathbf{w}_s^\top \mathbf u_{i,j} * \zeta_i, ~~
    \delta_{i,j} = \mathbf{w}_e^\top \mathbf u_{i,j} * \zeta_i
\end{align}

We extract the span with the highest span score $\gamma*\delta$ under the restriction that the start and end positions must belong to the same rule sentence.
Let $s$ and $e$ be the ground truth start and end position of the span. The underspecified span extraction loss is computed as the pointing loss
\begin{align}
    \mathcal{L}_{\text{span,s}} &= - \mathbbm{1}_{l=\text{inquire}} \log~\text{softmax}(\gamma)_s \\
    \mathcal{L}_{\text{span,e}} &= - \mathbbm{1}_{l=\text{inquire}} \log~\text{softmax}(\delta)_e
\end{align}

\noindent The overall loss is the sum of the decision loss, entailment prediction loss and span extraction loss
\begin{align}
    \mathcal{L} = \mathcal{L}_{\text{dec}} + \lambda_1 \mathcal{L}_{\text{entail}} + \lambda_2 \mathcal{L}_{\text{span}} \label{eq:totalloss}
\end{align}
where $\lambda_1$ and $\lambda_2$ are tunable hyperparameters.

\paragraph{Question Rephrasing.}
The underspecified span extracted in the previous stage is fed into the question rephrasing model to generate a follow-up question. We finetune the UniLM \cite{unilm} to achieve this goal. UniLM is a pretrained language model which demonstrates its effectiveness in both natural language understanding and generation tasks. Specifically, it outperforms previous methods by a large margin on the SQuAD question generation task \cite{du-cardie-2018-harvesting}.

As shown in Figure \ref{fig:model}, UniLM takes the concatenation of rule text and the extracted rule span as input, separated by the sentinel tokens:\noindent\texttt{[CLS]} rule-text \texttt{[SEP]} extracted-span \texttt{[SEP]}. 
The training target is the follow-up question we want to generate. Please refer to \newcite{unilm} for details on finetuning UniLM and doing inference with it.

\section{Experiments}

\begin{table*}[!t]
    \centering
    \small 
    \begin{tabular}{@{~~~}l@{~} | @{~}c@{~} @{~}c@{~}  @{~~~~~}c@{~~~~} @{~~~~}c@{~~~~~} } 
    \Xhline{2\arrayrulewidth}
    \multirow{2}{*}{Models} &  \multicolumn{4}{c}{End-to-End Task (Leaderboard Performance)}  \\  
     & Micro Acc. & Macro Acc. & BLEU1 & BLEU4  \\
    \hline\hline
    Seq2Seq \cite{saeidi-etal-2018-interpretation} & 44.8 & 42.8 & 34.0 & {~}{~}7.8   \\
    Pipeline \cite{saeidi-etal-2018-interpretation}  & 61.9 & 68.9 & 54.4 & 34.4   \\
    BERTQA \cite{zhong-zettlemoyer-2019-e3}  & 63.6 & 70.8 & 46.2 & 36.3  \\
    UrcaNet \cite{Sharma2019NeuralCQ}   & 65.1 & 71.2 & 60.5 & 46.1   \\
    BiSon \cite{lawrence-etal-2019-attending}  & 66.9 & 71.6 & 58.8 & 44.3  \\
    E${}^{3}$ \cite{zhong-zettlemoyer-2019-e3}  & 67.6 & 73.3 & 54.1 & 38.7   \\
    EMT (our single model)  & \textbf{69.1} & \textbf{74.6} & \textbf{63.9} & \textbf{49.5}  \\
    \Xhline{2\arrayrulewidth}
    \end{tabular}
    \caption{
    Performance on the blind, held-out test set of ShARC end-to-end task.
    }
    \label{tab:result-test}
\end{table*}

\subsection{Experimental Setup}\label{sec:ex_setup}
\paragraph{Dataset.}

We conduct experiments on the ShARC CMR dataset \cite{saeidi-etal-2018-interpretation}.
It contains 948 dialog trees, which are flattened into 32,436 examples by considering all possible nodes in the trees.
Each example is a quintuple of (rule text, initial question, user scenario, dialog history, decision), where decision is either one of \{\texttt{Yes}, \texttt{No}, \texttt{Irrelevant}\} or a follow-up question. The train, development, and test dataset sizes are 21890, 2270, and 8276, respectively.\footnote{Leaderboard: \url{https://sharc-data.github.io/leaderboard.html}}

\paragraph{End-to-End Evaluation.}
Organizers of the ShARC competition evaluate model performance as an end-to-end task.
They first evaluate the micro- and macro-accuracy for the decision making task.
If both the ground truth decision and the predicted decision are \texttt{Inquire}, then they evaluate the generated follow-up question using BLEU  score \cite{papineni-etal-2002-bleu}. However, this way of evaluating follow-up questions has one issue. If two models have different \texttt{Inquire} predictions, the follow-up questions for evaluation will be different, making the comparison unfair. For example, a model could classify only one example as \texttt{Inquire} in the whole test set and generate the follow-up question correctly, achieving a 100\% BLEU score. Therefore, we also propose to evaluate the follow-up question generation performance in an oracle evaluation setup as described below.

\paragraph{Oracle Question Generation Evaluation.}
In this evaluation, we ask the models to generate follow-up questions whenever the ground truth decision is \texttt{Inquire}, and compute the BLEU score for the generated questions accordingly. In this setup, there are 6804 examples for training and 562 examples for evaluation. 

\paragraph{Data Augmentation.}
In the annotation process of the ShARC dataset, the \textit{scenario} is manually constructed from a part of the dialog history, and that excerpt
 of the dialog is not shown as input to the model. Instead, it is treated as the \textit{evidence} which should be entailed from the scenario. 
To effectively utilize this additional signal, we construct more examples by replacing the \textit{scenario} with the \textit{evidence}. This leads to additional 5800 training instances.
We use this augmented dataset for the EMT model and its ablations in our experiments.

\paragraph{Labeling Underspecified Spans.}
To supervise the process of coarse-to-fine reasoning, we follow \newcite{zhong-zettlemoyer-2019-e3} to label the rule spans.
We first trim the follow-up questions in the conversation by removing question words ``do, does, did, is, was, are, have'' and the question mark ``?''.
For each trimmed question, we find the shortest span inside the rule text which has the minimum edit distance from the trimmed question, and treat it as an \texttt{underspecified span}.

\paragraph{Acquiring Labels for Entailment.}
To supervise the subtask of entailment prediction for each rule sentence, we use a heuristic to automatically label its  entailment state. For each rule sentence, we first find if it contains any {underspecified} span for the questions in the dialog history (and evidence text), and use the corresponding \texttt{Yes/No} answers to label the rule text as \texttt{Entailment}/\texttt{Contradiction}. The rule text without any underspecified span is labeled as \texttt{Unknown}. 


\paragraph{Implementation Details.}
We tokenize all text inputs with spaCy \cite{spacy2}.
The EMT model and the follow-up question generation model UniLM are trained separately and pipelined together at test time.
For EMT, we use the uncased BERT base model \cite{Wolf2019HuggingFacesTS} for encoding.
We train EMT with Adam \cite{adam} optimizer with a learning rate of 5e-5, a warm-up rate of 0.1 and a dropout rate of 0.35. 
The loss weights $\lambda_1$ and $\lambda_2$ in Eq. \ref{eq:totalloss} are set to 10 and 0.6 respectively, based on the development set results.
For UniLM, we fine-tuning it with a batch size of 16 and a learning rate of 2e-5, and we use a beam size of 10 for inference.

To reduce the variance of our experimental results, all experiments reported on the development set are repeated 5 times with different random seeds. We report the average results along with their standard deviations.

\begin{table}[!t]
    \centering
    \resizebox{1.0\columnwidth}{!}{
    \begin{tabular}{l | @{~~~~~~~}c@{~~~~~~~~~} @{~~~~~~}c@{~~~~~~~~~~}  c c }
    \Xhline{2\arrayrulewidth}
  Models & \texttt{Yes} & \texttt{No} & \texttt{Inquire} & \texttt{Irrelevant}  \\
    \hline
    \hline
    BERTQA    & 61.2 & 61.0  &  62.6 & 96.4       \\
    E${}^{3}$ & 65.9 & 70.6 & 60.5 & 96.4   \\
    UrcaNet* & 63.3 & 68.4 & 58.9 & 95.7   \\
    EMT   & \textbf{70.5} & \textbf{73.2} & \textbf{70.8} & \textbf{98.6}  \\
    \Xhline{2\arrayrulewidth}
    \end{tabular}}
    \caption{
    Class-wise decision prediction accuracy on the development set (*: reported in the paper).
    }
    \label{tab:result-clf}
\end{table}

\begin{table}[!t]
    \small
    \centering
    \resizebox{1.0\columnwidth}{!}{
    \begin{tabular}{l | c c | c c }
    \Xhline{2\arrayrulewidth}
    \multirow{3}{*}{Models}  & \multicolumn{4}{c}{Oracle Question Generation Task}  \\  
    & \multicolumn{2}{c}{Development Set} & \multicolumn{2}{c}{Cross Validation}  \\  
   & BLEU1 & \multicolumn{1}{c}{BLEU4} & BLEU1 & BLEU4  \\
    \hline
    \hline
    E${}^{3}$   & 52.79\scriptsize{$\pm$2.87} & 37.31\scriptsize{$\pm$2.35} & 51.75 & 35.94    \\
    E${}^{3}$+UniLM & 57.09\scriptsize{$\pm$1.70} & 41.05\scriptsize{$\pm$1.80} & 56.94 & 42.87      \\
    EMT  & \textbf{62.32}\scriptsize{$\pm$1.62} & \textbf{47.89}\scriptsize{$\pm$1.58} & \textbf{64.48} & \textbf{52.40}  \\
    \Xhline{2\arrayrulewidth}
    \end{tabular}}
    \caption{
    Performance on Oracle Question Generation Task. We show both results on the development set and 10-fold cross validation. 
    E${}^{3}$+UniLM replaces the editor of E${}^{3}$ to our finetuned UniLM.
    }
    \label{tab:result-qg}
\end{table}

\subsection{Results}
\begin{table*}[!t]
    \small
    \centering
    \begin{tabular}{l | c c c c | c c }
    \Xhline{2\arrayrulewidth}
    & \multicolumn{4}{c|}{End-to-End Task} & \multicolumn{2}{c}{Oracle Question Generation Task}  \\  
    Models & Micro Acc. & Macro Acc. & BLEU1 & BLEU4 & ~~~~BLEU1 & BLEU4  \\
    \hline
    \hline
    EMT   & \textbf{71.36}\scriptsize{$\pm$0.69} & \textbf{76.70}\scriptsize{$\pm$0.54} & \textbf{67.04}\scriptsize{$\pm$1.59} & \textbf{52.37}\scriptsize{$\pm$1.92} & ~~~~\textbf{63.53}\scriptsize{$\pm$1.03} & \textbf{48.69}\scriptsize{$\pm$0.80}     \\
    EMT (w/o data aug.)  & 70.67\scriptsize{$\pm$0.52} & 76.33\scriptsize{$\pm$0.69} & 65.86\scriptsize{$\pm$2.25} & 51.02\scriptsize{$\pm$2.52} & ~~~~62.38\scriptsize{$\pm$1.34} & 47.58\scriptsize{$\pm$1.30}     \\
    EMT (w/o c2f)  & 70.41\scriptsize{$\pm$0.94} & 75.96\scriptsize{$\pm$0.91} & 65.73\scriptsize{$\pm$1.76} & 50.84\scriptsize{$\pm$2.31} & ~~~~61.98\scriptsize{$\pm$1.26} & 47.66\scriptsize{$\pm$1.33}     \\
    EMT (w/o $\mathcal{L}_{\text{entail}}$)  & 67.81\scriptsize{$\pm$1.20} & 73.50\scriptsize{$\pm$0.83} & 63.84\scriptsize{$\pm$1.80} & 49.35\scriptsize{$\pm$2.10} & ~~~~60.50\scriptsize{$\pm$1.16} & 45.34\scriptsize{$\pm$1.73}  \\
    EMT (w/o tracker)  & 67.42\scriptsize{$\pm$1.15} & 72.73\scriptsize{$\pm$0.74} & 63.26\scriptsize{$\pm$0.64} & 47.97\scriptsize{$\pm$0.40} & ~~~~61.87\scriptsize{$\pm$1.46} & 47.13\scriptsize{$\pm$1.35}  \\
    \Xhline{2\arrayrulewidth}
    \end{tabular}
    \caption{
    Ablation Study of EMT on the development set of ShARC. 
    }
    \label{tab:result-ablation}
\end{table*}

\paragraph{End-to-End Task.}
The end-to-end performance on the held-out test set is shown in Table \ref{tab:result-test}.
EMT outperforms the existing state-of-the-art model E${}^{3}$ on decision classification in both micro- and macro-accuracy.
Although the BLEU scores are not directly comparable among different models, EMT achieves competitive BLEU1 and BLEU4 scores on the examples it makes an \texttt{Inquire} decision.
The results show that EMT has strong capability in both decision making and follow-up question generation tasks.
Table \ref{tab:result-clf} presents the  class-wise accuracy on the four decision types.
EMT improves on the  \texttt{Inquire} decision significantly. It is because EMT  can explicitly track the states of all rule sentences; it has a macro accuracy of 80\%  on the entailment state prediction task. 

\paragraph{Oracle Question Generation Task.}
To establish a concrete question generation evaluation, we conduct experiments on our proposed oracle question generation task.
We compare our model EMT with E${}^{3}$ and an extension E${}^{3}$+UniLM; implementations for other methods are not publicly available.
E${}^{3}$+UniLM replaces the editor of E${}^{3}$ with our finetuned UniLM.
The results on the development set and 10-fold cross validation are shown in Table \ref{tab:result-qg}.

Firstly, E${}^{3}$+UniLM performs better than E${}^{3}$, validating the effectiveness of our follow-up question rephrasing module: finetuned UniLM.
More importantly, EMT consistently outperforms  E${}^{3}$ and E${}^{3}$+UniLM on both the development set and the cross validation by a large margin.
Although there is no ground truth label for span extraction, we can infer from the question generation results that our coarse-to-fine reasoning approach extracts better spans than the extraction and retrieval modules of E${}^{3}$.
This is because E${}^{3}$ propagates error from the span extraction module to the span retrieval module while our coarse-to-fine approach avoids this problem through weighting token-level span distributions with its sentence-level entailment scores.

\subsection{Ablation Study}\label{sec:ablation}
We conduct an ablation study on the development set for both the end-to-end evaluation task and oracle question generation evaluation task.
We consider four ablations of our EMT model:
\begin{itemize}

\item[(1)] EMT (w/o data aug.) trains the model on the original ShARC training set and do not use any augmented data using the evidence.\vspace{-0.1in}
\item[(2)] EMT (w/o c2f) extracts the rule span without weighted by the entailment score $\zeta$ in Eqn. \ref{eqn:modulate}.\vspace{-0.1in}
\item[(3)] EMT (w/o $\mathcal{L}_{\text{entail}}$) removes the entailment state prediction subtask in decision making, and thus there is no entailment score $\zeta$ for underspecified span extraction in Eqn. \ref{eqn:modulate}.\vspace{-0.1in}
\item[(4)] EMT (w/o tracker) that removes the explicit memory tracking module. Instead, it treats the \texttt{[CLS]} token for each rule sentence as the state for decision making and span extraction.
\end{itemize}

\begin{figure*}[t!]
\centering
\includegraphics[width=1.0\textwidth]{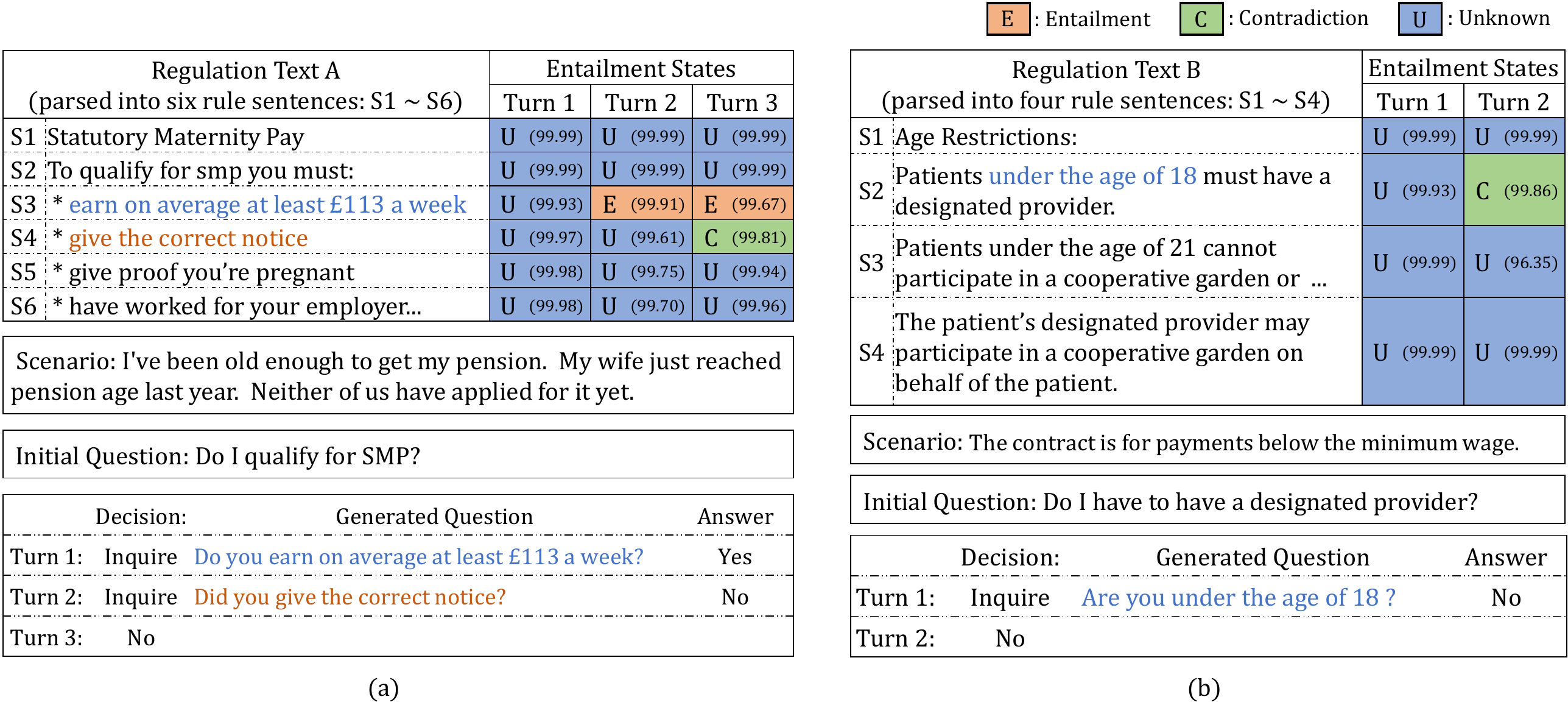}
\caption{Predicted decisions and generated questions by our EMT model. Extracted spans and their corresponding questions are marked in the same colors.
We also visualize the transitions of predicted entailment states (\textbf{E}ntailment, {\textbf{C}ontradiction}, {\textbf{U}nknown}) over rule sentences (S1, S2, S3 ...) as the conversation flows, with associated entailment scores [$\beta_{\text{entailment}}$, $\beta_{\text{contradiction}}$, $\beta_{\text{unknown}}$].
}
\label{fig:interpret}
\end{figure*}

Results of the ablations are shown in Table \ref{tab:result-ablation}, and we have the following observations:
\begin{itemize}[wide=0\parindent]
    \item With the help of data augmentation, EMT boosts the performance slightly on the end-to-end task, especially for the question generation task which originally has only 6804 training examples. The augmented training instances boosts the performance even though the augmentation method does not produce any new question. This implies that the size of the ShARC dataset is a bottleneck for an effective end-to-end neural models.  
    
    \item Without the coarse-to-fine reasoning for span extraction, EMT (w/o c2f) drops by 1.53 on BLEU4, which implies that it is necessary for the question generation task.
    The reason is that, as a classification task, entailment state prediction can be trained reasonably well (80\% macro accuracy) with a limited amount of data (6804 training examples). 
    Therefore, the \texttt{Unknown} scores in the entailment state prediction can guide the span extraction via a soft modulation (Equation \ref{eqn:modulate}).
    On the other hand, one-step span extraction method does not utilize the entailment states of the rule sentences from EMT, meaning it does not learn to extract the underspecified part of the rule text.
    \item With the guidance of explicit entailment supervision, EMT outperforms EMT (w/o $\mathcal{L}_{\text{entail}}$) by a large margin. Intuitively, knowing the entailment states of the rule sentences makes the decision making process easier for complex tasks that require logical reasoning on conjunctions of conditions or disjunctions of conditions. It also helps span extraction through the coarse-to-fine approach.

    \item Without the explicit memory tracker described in Section \ref{sec:emt}, EMT (w/o tracker) performs poorly on the decision making task. Although there exist interactions between rule sentences and user information in BERT-encoded representations through multi-head self-attentions, it is not adequate to learn whether conditions listed in the rule text have already been satisfied or not.
\end{itemize}

\subsection{Interpretability}
To get better insights into the underlying entailment-oriented reasoning process of EMT, we examine the entailment states of the rule sentences as the conversation flows. Two example cases are provided in Figure \ref{fig:interpret}.
Given a rule text containing several rule sentences (S1, S2, S3, ...), we show the transition of predicted entailment states [$\beta_{\text{entailment}}$, $\beta_{\text{contradiction}}$, $\beta_{\text{unknown}}$] over multiple turns in the dialogue.

\paragraph{Rules in Bullet Points.}
Figure \ref{fig:interpret} (a) shows an example in which the rule text is expressed in the conjunction of four bullet-point conditions.
On the first turn, EMT reads ``Scenario'' and ``Initial Question'' and they only imply that the question from the user is relevant to the rule text.
Thus the entailment states for all the rule sentences are \texttt{Unknown}, and EMT makes an  \texttt{Inquire} decision, and asks a question. 
Once a positive answer is received from the user part for the first turn, EMT transits the entailment state for rule sentence S3 from \texttt{Unknown} to \texttt{Entailment}, but it still cannot conclude the dialogue, so it asks a second follow-up question. Then we see that the user response for the second question is negative, which makes EMT conclude a final decision \texttt{No} in the third turn.

\paragraph{Rules in Plain Text.}
Figure \ref{fig:interpret} (b) presents a more challenging case where the rules are in plain text.
Therefore, it is not possible to put the whole sentence into a clarification question as EMT in Figure \ref{fig:interpret}(a) does.
In this case, both the decision making module and span extraction module contribute to helping the user.
The span extraction module locates the correct spans inside S2, and EMT concludes a correct answer ``No'' after knowing the user does not fulfill the condition listed in S2.

\subsection{Error Analysis}
We analyze some errors of EMT predictions on the ShARC development set, as described below.
\paragraph{Decision Making Error.}
Out of 2270 examples in the development set, our EMT produces incorrect decisions on 608 cases.
We manually analyze 104 error cases.
In 40 of these cases, EMT fails to derive the correct entailment states for each rule sentence, while in 23 cases, the model predicts the correct entailment states but cannot predict correct decisions based on that. These errors suggest that explicitly modeling the logic reasoning process is a promising direction.
Another challenge comes from extracting useful information from the user scenarios. In 24 cases, the model fails to make the correct decision because it could not infer necessary user information from the scenarios.
Last but not least, parsing the rule text into rule sentences is also a challenge. As shown in Figure \ref{fig:interpret}(b), the plain text usually contains complicated clauses for rule conditions, which is difficult to disentangle them into separate conditions. In 17 cases, one single rule sentence contains multiple conditions, which makes the model fail to conduct the entailment reasoning correctly.

\paragraph{Question Generation Error.}
Out of 562 question generation examples in the development set, our EMT locates the underspecified span poorly in 115 cases (span extraction F1 score $\leq$ 0.5).
We manually analyze 52 wrong question generation cases.
Out of 29 cases of them, EMT fails to predict correct entailment states for rule sentences, and thus does not locate the span within the ground truth rule sentence, while in 9 cases, it finds the correct rule sentence but extracts a different span.
Another challenge comes from the one-to-many problem in sequence generation. When there are multiple underspecified rule sentences, the model asks about one of these underspecified rule sentences which is different from the ground truth one. This suggests that new evaluation metrics could be proposed by taking this into consideration.

\section{Related Work}
\paragraph{ShARC Conversational Machine Reading} \cite{saeidi-etal-2018-interpretation} differs from conversational question answering \cite{choi-etal-2018-quac,reddy-etal-2019-coqa} and conversational question generation \cite{gao-etal-2019-interconnected} in that 1) machines are required to formulate follow-up questions to fill the information gap, and 2) machines have to interpret a set of complex decision rules and make a question-related conclusion, instead of extracting the answer from the text. CMR can be viewed as a special type of task-oriented dialog systems \cite{wen-etal-2017-network,zhong-etal-2018-global,wu-etal-2019-transferable} to help users achieve their goals. However, it does not rely on predefined slot and ontology information but natural language rules.

On the ShARC CMR challenge \cite{saeidi-etal-2018-interpretation}, \citet{lawrence-etal-2019-attending} propose an end-to-end bidirectional sequence generation approach with mixed decision making and question generation stages. \citet{saeidi-etal-2018-interpretation} split it into sub-tasks and combines hand-designed sub-models for decision classification, entailment and question generation. \citet{zhong-zettlemoyer-2019-e3} propose to extract all possible rule text spans, assign each of them an entailment score, and edit the span with the highest score into a follow-up question. However, they do not use these entailment scores for decision making. \citet{Sharma2019NeuralCQ} study patterns of the dataset and include additional embeddings from dialog history and user scenario as rule markers to help decision making. Compared to these methods, our EMT has two key differences: (1) EMT makes decision via explicitly entailment-oriented reasoning, which, to our knowledge, is the first such approach; (2) Instead of treating decision making and follow-up question generation (or span extraction) separately, EMT is a unified approach that exploits its memory states for both decision making and question generation.

\vspace{-0.05in}
\paragraph{Memory-Augmented Neural Networks.}
Our work is also related to memory-augmented neural networks~\cite{graves2014neural,graves2016hybrid}, which have been applied in some NLP tasks such as question answering~\cite{DBLP:conf/iclr/HenaffWSBL17} and machine translation~\cite{wang-etal-2016-memory}. For dialog applications, \citet{zhang2019memory} propose a dialogue management model that employs a memory controller and a slot-value memory, \citet{bordes2016learning} learn a restaurant bot by end-to-end memory networks, \citet{madotto-etal-2018-mem2seq} incorporate external memory modules into dialog generation.

\section{Conclusions}

In this paper, we have proposed a new framework for conversational machine reading (CMR) that comprises a novel explicit memory tracker (EMT) to track entailment states of the rule sentences explicitly within its memory module. The updated states are utilized for decision making and coarse-to-fine follow-up question generation in a unified manner. EMT achieved a new state-of-the-art result on the ShARC CMR challenge. EMT also gives interpretability by showing the entailment-oriented reasoning process as the conversation flows. 
While we conducted experiments on the ShARC dataset, we believe the proposed methodology could be extended to other kinds of CMR tasks. 

\section*{Acknowledgments}

We thank Max Bartolo and Patrick Lewis for evaluating our submitted models on the hidden test set, and for their helpful replies on dataset related questions. 
Also thanks to Victor Zhong for replies on replicating the E$^3$ model.
The work described in this paper was partially supported by following projects from the Research Grants Council of the Hong Kong Special Administrative Region, China: CUHK 2300174 (Collaborative Research Fund, No. C5026-18GF); CUHK 14210717 (RGC General Research Fund).

\bibliography{acl2020}
\bibliographystyle{acl_natbib}

\appendix
\newpage
\section{Appendices}
\input{final_appendix.tex}

\end{document}

%% file: final_appendix.tex
In our preliminary investigation, we did not add the entailment state prediction subtask (Eqn.\ref{eqn:score_entail} \& \ref{eqn:l_entail}) in Section \ref{sec:clf}. Consequently, there is no sentence-level entailment score in Eqn.\ref{eqn:entail_sent_score} for coarse-to-fine reasoning.
Instead, we tried to predict the underspecified rule sentence separately and treat it as the sentence-level score:
\begin{align}
    \zeta_i = \mathbf w_\zeta [\mathbf k_i; \mathbf v_{i}] + b_\zeta \in \real{}
\end{align}
where $\zeta_i$ is the score that determines how likely the $i$-th rule sentence contains the underspecified span. Let $l$ indicate the correct decision and $j$ be the underspecified rule sentence, the identification loss of the underspecified rule sentence is
\begin{align}
    \mathcal{L}_{\text{sent}} = - \mathbbm{1}_{l=\text{inquire}} \log~\text{softmax}(\zeta)_j
\end{align}

On the ShARC hidden test set, it turns out that this model EMT performs slightly better on the decision making part but worse than EMT+entailment (the model described in the body text) for the question generation task. However, it is hard to balance the decision making performance and the question generation (span extraction) performance for this model EMT, and thus we develop the current version described in the main paper which introduces an entailment state prediction subtask.
Table \ref{tab:emt-v0} shows the results of these two models on the ShARC test set.

\begin{table}[h]
    \centering
    \small 
    \begin{tabular}{@{~~~}l@{~} | @{~}c@{~} @{~}c@{~}  @{~}c@{~} @{~}c@{~} } 
    \Xhline{2\arrayrulewidth}
    \multirow{2}{*}{Models} &  \multicolumn{4}{c}{End-to-End Task}  \\  
     & Micro Acc. & Macro Acc. & BLEU1 & BLEU4  \\
    \hline\hline
    EMT & \textbf{69.4} & \textbf{74.8} & {60.9} & {46.0}  \\
    EMT+entailment  & {69.1} & {74.6} & \textbf{63.9} & \textbf{49.5}  \\
    \Xhline{2\arrayrulewidth}
    \end{tabular}
    \caption{
    Results of EMT+entailment and EMT on the hidden test set of ShARC.
    }
    \label{tab:emt-v0}
\end{table}

%% file: acl2020.bbl
\begin{thebibliography}{25}
\expandafter\ifx\csname natexlab\endcsname\relax\def\natexlab#1{#1}\fi

\bibitem[{Bordes et~al.(2016)Bordes, Boureau, and Weston}]{bordes2016learning}
Antoine Bordes, Y-Lan Boureau, and Jason Weston. 2016.
\newblock Learning end-to-end goal-oriented dialog.
\newblock \emph{arXiv preprint arXiv:1605.07683}.

\bibitem[{Choi et~al.(2018)Choi, He, Iyyer, Yatskar, Yih, Choi, Liang, and
  Zettlemoyer}]{choi-etal-2018-quac}
Eunsol Choi, He~He, Mohit Iyyer, Mark Yatskar, Wen-tau Yih, Yejin Choi, Percy
  Liang, and Luke Zettlemoyer. 2018.
\newblock \href {https://doi.org/10.18653/v1/D18-1241} {{Q}u{AC}: Question
  answering in context}.
\newblock In \emph{Proceedings of the 2018 Conference on Empirical Methods in
  Natural Language Processing}, pages 2174--2184, Brussels, Belgium.
  Association for Computational Linguistics.

\bibitem[{Devlin et~al.(2019)Devlin, Chang, Lee, and
  Toutanova}]{devlin-etal-2019-bert}
Jacob Devlin, Ming-Wei Chang, Kenton Lee, and Kristina Toutanova. 2019.
\newblock \href {https://doi.org/10.18653/v1/N19-1423} {{BERT}: Pre-training of
  deep bidirectional transformers for language understanding}.
\newblock In \emph{Proceedings of the 2019 Conference of the North {A}merican
  Chapter of the Association for Computational Linguistics: Human Language
  Technologies, Volume 1 (Long and Short Papers)}, pages 4171--4186,
  Minneapolis, Minnesota. Association for Computational Linguistics.

\bibitem[{Dong et~al.(2019)Dong, Yang, Wang, Wei, Liu, Wang, Gao, Zhou, and
  Hon}]{unilm}
Li~Dong, Nan Yang, Wenhui Wang, Furu Wei, Xiaodong Liu, Yu~Wang, Jianfeng Gao,
  Ming Zhou, and Hsiao-Wuen Hon. 2019.
\newblock Unified language model pre-training for natural language
  understanding and generation.
\newblock In \emph{33rd Conference on Neural Information Processing Systems
  (NeurIPS 2019)}.

\bibitem[{Du and Cardie(2018)}]{du-cardie-2018-harvesting}
Xinya Du and Claire Cardie. 2018.
\newblock \href {https://doi.org/10.18653/v1/P18-1177} {Harvesting
  paragraph-level question-answer pairs from {W}ikipedia}.
\newblock In \emph{Proceedings of the 56th Annual Meeting of the Association
  for Computational Linguistics (Volume 1: Long Papers)}, pages 1907--1917,
  Melbourne, Australia. Association for Computational Linguistics.

\bibitem[{Gao et~al.(2019)Gao, Li, King, and
  Lyu}]{gao-etal-2019-interconnected}
Yifan Gao, Piji Li, Irwin King, and Michael~R. Lyu. 2019.
\newblock \href {https://doi.org/10.18653/v1/P19-1480} {Interconnected question
  generation with coreference alignment and conversation flow modeling}.
\newblock In \emph{Proceedings of the 57th Annual Meeting of the Association
  for Computational Linguistics}, pages 4853--4862, Florence, Italy.
  Association for Computational Linguistics.

\bibitem[{Graves et~al.(2014)Graves, Wayne, and Danihelka}]{graves2014neural}
Alex Graves, Greg Wayne, and Ivo Danihelka. 2014.
\newblock Neural turing machines.
\newblock \emph{arXiv preprint arXiv:1410.5401}.

\bibitem[{Graves et~al.(2016)Graves, Wayne, Reynolds, Harley, Danihelka,
  Grabska-Barwi{\'n}ska, Colmenarejo, Grefenstette, Ramalho, Agapiou
  et~al.}]{graves2016hybrid}
Alex Graves, Greg Wayne, Malcolm Reynolds, Tim Harley, Ivo Danihelka, Agnieszka
  Grabska-Barwi{\'n}ska, Sergio~G{\'o}mez Colmenarejo, Edward Grefenstette,
  Tiago Ramalho, John Agapiou, et~al. 2016.
\newblock Hybrid computing using a neural network with dynamic external memory.
\newblock \emph{Nature}, 538(7626):471.

\bibitem[{Henaff et~al.(2017)Henaff, Weston, Szlam, Bordes, and
  LeCun}]{DBLP:conf/iclr/HenaffWSBL17}
Mikael Henaff, Jason Weston, Arthur Szlam, Antoine Bordes, and Yann LeCun.
  2017.
\newblock \href {https://openreview.net/forum?id=rJTKKKqeg} {Tracking the world
  state with recurrent entity networks}.
\newblock In \emph{5th International Conference on Learning Representations,
  {ICLR} 2017, Toulon, France, April 24-26, 2017, Conference Track
  Proceedings}. OpenReview.net.

\bibitem[{Honnibal and Montani(2017)}]{spacy2}
Matthew Honnibal and Ines Montani. 2017.
\newblock {spaCy 2}: Natural language understanding with {B}loom embeddings,
  convolutional neural networks and incremental parsing.
\newblock To appear.

\bibitem[{Kingma and Ba(2015)}]{adam}
Diederik~P. Kingma and Jimmy Ba. 2015.
\newblock \href {http://arxiv.org/abs/1412.6980} {Adam: {A} method for
  stochastic optimization}.
\newblock In \emph{3rd International Conference on Learning Representations,
  {ICLR} 2015, San Diego, CA, USA, May 7-9, 2015, Conference Track
  Proceedings}.

\bibitem[{Lawrence et~al.(2019)Lawrence, Kotnis, and
  Niepert}]{lawrence-etal-2019-attending}
Carolin Lawrence, Bhushan Kotnis, and Mathias Niepert. 2019.
\newblock \href {https://doi.org/10.18653/v1/D19-1001} {Attending to future
  tokens for bidirectional sequence generation}.
\newblock In \emph{Proceedings of the 2019 Conference on Empirical Methods in
  Natural Language Processing and the 9th International Joint Conference on
  Natural Language Processing (EMNLP-IJCNLP)}, pages 1--10, Hong Kong, China.
  Association for Computational Linguistics.

\bibitem[{Madotto et~al.(2018)Madotto, Wu, and
  Fung}]{madotto-etal-2018-mem2seq}
Andrea Madotto, Chien-Sheng Wu, and Pascale Fung. 2018.
\newblock \href {https://doi.org/10.18653/v1/P18-1136} {{M}em2{S}eq:
  Effectively incorporating knowledge bases into end-to-end task-oriented
  dialog systems}.
\newblock In \emph{Proceedings of the 56th Annual Meeting of the Association
  for Computational Linguistics (Volume 1: Long Papers)}, pages 1468--1478,
  Melbourne, Australia. Association for Computational Linguistics.

\bibitem[{Papineni et~al.(2002)Papineni, Roukos, Ward, and
  Zhu}]{papineni-etal-2002-bleu}
Kishore Papineni, Salim Roukos, Todd Ward, and Wei-Jing Zhu. 2002.
\newblock \href {https://doi.org/10.3115/1073083.1073135} {{B}leu: a method for
  automatic evaluation of machine translation}.
\newblock In \emph{Proceedings of the 40th Annual Meeting of the Association
  for Computational Linguistics}, pages 311--318, Philadelphia, Pennsylvania,
  USA. Association for Computational Linguistics.

\bibitem[{Reddy et~al.(2019)Reddy, Chen, and Manning}]{reddy-etal-2019-coqa}
Siva Reddy, Danqi Chen, and Christopher~D. Manning. 2019.
\newblock \href {https://doi.org/10.1162/tacl_a_00266} {{C}o{QA}: A
  conversational question answering challenge}.
\newblock \emph{Transactions of the Association for Computational Linguistics},
  7:249--266.

\bibitem[{Saeidi et~al.(2018)Saeidi, Bartolo, Lewis, Singh, Rockt{\"a}schel,
  Sheldon, Bouchard, and Riedel}]{saeidi-etal-2018-interpretation}
Marzieh Saeidi, Max Bartolo, Patrick Lewis, Sameer Singh, Tim Rockt{\"a}schel,
  Mike Sheldon, Guillaume Bouchard, and Sebastian Riedel. 2018.
\newblock \href {https://doi.org/10.18653/v1/D18-1233} {Interpretation of
  natural language rules in conversational machine reading}.
\newblock In \emph{Proceedings of the 2018 Conference on Empirical Methods in
  Natural Language Processing}, pages 2087--2097, Brussels, Belgium.
  Association for Computational Linguistics.

\bibitem[{Sharma et~al.(2019)Sharma, Contractor, Kumar, and
  Joshi}]{Sharma2019NeuralCQ}
Abhishek Sharma, Danish Contractor, Harshit Kumar, and Sachindra Joshi. 2019.
\newblock Neural conversational qa: Learning to reason v.s. exploiting
  patterns.
\newblock \emph{ArXiv}, abs/1909.03759.

\bibitem[{Vaswani et~al.(2017)Vaswani, Shazeer, Parmar, Uszkoreit, Jones,
  Gomez, Kaiser, and Polosukhin}]{transformer}
Ashish Vaswani, Noam Shazeer, Niki Parmar, Jakob Uszkoreit, Llion Jones,
  Aidan~N Gomez, \L~ukasz Kaiser, and Illia Polosukhin. 2017.
\newblock \href
  {http://papers.nips.cc/paper/7181-attention-is-all-you-need.pdf} {Attention
  is all you need}.
\newblock In \emph{Advances in Neural Information Processing Systems 30}, pages
  5998--6008. Curran Associates, Inc.

\bibitem[{Wang et~al.(2016)Wang, Lu, Li, and Liu}]{wang-etal-2016-memory}
Mingxuan Wang, Zhengdong Lu, Hang Li, and Qun Liu. 2016.
\newblock \href {https://doi.org/10.18653/v1/D16-1027} {Memory-enhanced decoder
  for neural machine translation}.
\newblock In \emph{Proceedings of the 2016 Conference on Empirical Methods in
  Natural Language Processing}, pages 278--286, Austin, Texas. Association for
  Computational Linguistics.

\bibitem[{Wen et~al.(2017)Wen, Vandyke, Mrk{\v{s}}i{\'c}, Ga{\v{s}}i{\'c},
  Rojas-Barahona, Su, Ultes, and Young}]{wen-etal-2017-network}
Tsung-Hsien Wen, David Vandyke, Nikola Mrk{\v{s}}i{\'c}, Milica
  Ga{\v{s}}i{\'c}, Lina~M. Rojas-Barahona, Pei-Hao Su, Stefan Ultes, and Steve
  Young. 2017.
\newblock \href {https://www.aclweb.org/anthology/E17-1042} {A network-based
  end-to-end trainable task-oriented dialogue system}.
\newblock In \emph{Proceedings of the 15th Conference of the {E}uropean Chapter
  of the Association for Computational Linguistics: Volume 1, Long Papers},
  pages 438--449, Valencia, Spain. Association for Computational Linguistics.

\bibitem[{Wolf et~al.(2019)Wolf, Debut, Sanh, Chaumond, Delangue, Moi, Cistac,
  Rault, Louf, Funtowicz, and Brew}]{Wolf2019HuggingFacesTS}
Thomas Wolf, Lysandre Debut, Victor Sanh, Julien Chaumond, Clement Delangue,
  Anthony Moi, Pierric Cistac, Tim Rault, R'emi Louf, Morgan Funtowicz, and
  Jamie Brew. 2019.
\newblock Huggingface's transformers: State-of-the-art natural language
  processing.
\newblock \emph{ArXiv}, abs/1910.03771.

\bibitem[{Wu et~al.(2019)Wu, Madotto, Hosseini-Asl, Xiong, Socher, and
  Fung}]{wu-etal-2019-transferable}
Chien-Sheng Wu, Andrea Madotto, Ehsan Hosseini-Asl, Caiming Xiong, Richard
  Socher, and Pascale Fung. 2019.
\newblock \href {https://doi.org/10.18653/v1/P19-1078} {Transferable
  multi-domain state generator for task-oriented dialogue systems}.
\newblock In \emph{Proceedings of the 57th Annual Meeting of the Association
  for Computational Linguistics}, pages 808--819, Florence, Italy. Association
  for Computational Linguistics.

\bibitem[{Zhang et~al.(2019)Zhang, Huang, Zhao, Ji, Chen, and
  Zhu}]{zhang2019memory}
Zheng Zhang, Minlie Huang, Zhongzhou Zhao, Feng Ji, Haiqing Chen, and Xiaoyan
  Zhu. 2019.
\newblock Memory-augmented dialogue management for task-oriented dialogue
  systems.
\newblock \emph{ACM Transactions on Information Systems (TOIS)}, 37(3):34.

\bibitem[{Zhong et~al.(2018)Zhong, Xiong, and Socher}]{zhong-etal-2018-global}
Victor Zhong, Caiming Xiong, and Richard Socher. 2018.
\newblock \href {https://doi.org/10.18653/v1/P18-1135} {Global-locally
  self-attentive encoder for dialogue state tracking}.
\newblock In \emph{Proceedings of the 56th Annual Meeting of the Association
  for Computational Linguistics (Volume 1: Long Papers)}, pages 1458--1467,
  Melbourne, Australia. Association for Computational Linguistics.

\bibitem[{Zhong and Zettlemoyer(2019)}]{zhong-zettlemoyer-2019-e3}
Victor Zhong and Luke Zettlemoyer. 2019.
\newblock \href {https://doi.org/10.18653/v1/P19-1223} {{E}3: Entailment-driven
  extracting and editing for conversational machine reading}.
\newblock In \emph{Proceedings of the 57th Annual Meeting of the Association
  for Computational Linguistics}, pages 2310--2320, Florence, Italy.
  Association for Computational Linguistics.

\end{thebibliography}
